# A Hybrid Framework for Action Recognition in Low-Quality Video Sequences

Tej Singh[1], Dinesh Kumar Vishwakarma[2]

[1]Department of Electronics and Communication Engineering, [2]Department of Information Technology, Delhi Technological University, Bawana Road, Delhi

[1]ttomar07@gmail.com, [2]dvishwakarma@gmail.com

**Abstract:** Vision-based activity recognition is essential for security, monitoring and surveillance applications. Further, real-time analysis having low-quality video and contain less information about surrounding due to poor illumination, and occlusions. Therefore, it needs a more robust and integrated model for low quality and night security operations. In this context, we proposed a hybrid model for illumination invariant human activity recognition based on sub-image histogram equalization enhancement and k-key pose human silhouettes. This feature vector gives good average recognition accuracy on three low exposure video sequences subset of original actions video datasets. Finally, the performance of the proposed approach is tested over three manually downgraded low qualities Weizmann action, KTH, and Ballet Movement dataset. This model outperformed on low exposure videos over existing technique and achieved comparable classification accuracy to similar state-of-the-art methods.

## 1 Introduction

Today human activity recognition in videos become a trendy area in computer vision community due to its wide range of applications such as video surveillance, robotics, patients monitoring, online gaming, sports, terrorist activities, content-based video analysis, and gait analysis [1] [2] [3] [4]. The major challenges in activity recognition in the video are varying illumination conditions, view variation, camera jitters, low resolution and cluttered background. At present most of the activity recognition solutions focused on high definition (HD) videos, but this is not suitable for low-quality video and many real-time video applications. That is why activity classification in low-contrast frames is one of the unexplored areas till now. It is difficult to extract the moving person from a dark background in a low contrast video [5] [6]. Further, it is necessary to enhance the quality of low contrast videos received for real-life applications. The video enhancement techniques aim to provide better visual appearance from an input of low exposure video to high exposure video for automated video processing, object detection, segmentation and human activity recognition [7]. It can be observed that many approaches toward action recognition in videos have been developed on different benchmark but very few on low-resolution videos. Further, it is challenging task to enhance the low-quality videos for specific applications due to low contrast, high ISO as compared to signal to noise ratio (SNR), acquiring source, environmental artefacts, inter-frames coherence and storage limitations. Therefore, a robust Human Activity Recognition (HAR) model need to developed to overcome the challenges present in low contrast videos and to recognise action classes in videos. With the objective to investigate and provide a unique solution to this problem, we propose a joint utilisation of histogram equalization image enhancement technique and spatiotemporal key poses feature representation model for action recognition in low qualities video sequences. In our model first, we enhanced the quality of manually degraded video frames by utilized work presented in [8] and recognised the action classes in such improved video frame. This hybrid approach represents a robust model for illumination invariant activity recognition.

## 2 Related Work

Based on existing literature survey there are two type video enhancement techniques: Spatial and Frequency transform domain video enhancement. In spatial based approaches image, pixels are directly manipulated while frequency based working on transform image obtained from the spatial spectrum. Till now various approaches [9] [10] [11] for video enhancement are proposed, but there is no unified solution which can apply for design criteria. The spatial domain based approaches are easy to implement with fewer time complexities and suitable for real-time applications. Further, these techniques high sensitive to camera jitters, and background changes [12] [10]. The video enhancement technique based on the transform domain operate on coefficients of images, e.g. Fourier transform [13] [11] [14]. These approaches have less computational complexities but difficult to synchronised the enhancement procedure on images simultaneously. Singh and Kapoor [8] proposed the low exposure enhancement technique, (ESIHE) based on sub-image histogram equalisation that improves the appearance quality of the image. The enhancement technique for low luminance images outperforms over other spatial histogram equalization technique. It is easy to extract the local and global texture features from an enhancement image so that images enhancement techniques have been useful for low-quality video frames.

The shape and motion feature descriptors methods are used to recognised human activity in video sequences. The shape feature is extracted from human body silhouettes while the motion feature relies on the optical flow of body motion. The shape descriptor is more useful as compared to motion because the motion feature is not so robust when the object is moving at different speeds. In work [15] proposed a silhouette based approach Motion History Images (MHI) and Motion Energy Images (MEI) template to recognised human activity. Chaaraoui et al. [16] introduced the human silhouettes based feature extraction approach and optimised these parameters for activity recognition. Thurau and Hlavac [17] introduced a Histogram of the Gradient-based approach using binary silhouettes to represent human action in video frames. Vishwakarma and Kapoor [18] proposed a hybrid classifier model for HAR system using key pose silhouettes. The key-frames are selected using maximum entropy energy frames compare with higher energy frames. See and Rheman [19] proposed a spatiotemporal based approach to recognise the action in low-quality videos. They test existing approached on down-sampled video of the original dataset to reduce the quality of videos. Similar methods were introduced in [20] [21] [22] on down-sampled video of low quality. The key ingredient of proposed work as follows:

- Low-quality input video sequences are enhanced using sub-histogram equalization technique.
- A robust spatiotemporal representation feature vector is obtained using key pose segmented human body silhouettes.
- A hybrid cascaded approach is developed through enhancement and representation of low-quality video sequences which is robust to illumination.
- Performance of the cascaded model is evaluated on three downsampled low exposure human action videos datasets, the recognition accuracy achieved on these dataset is compared with similar state-of-the-art methods and achieved superior performance.

The organisation of the proposed work as follows: Related work are discussed in section 2. Section 3 consists of a block diagram of the proposed model for human activity recognition. In section 4 the experimental setup and results are discussed on video dataset. The work is concluded in last section 5.

### 3 Proposed Framework of Activity Recognition

Our HAR model is based on the selection of k- key poses from enhancing output obtained from low-quality input video sequences using a low exposure refinement sub histogram equalization technique [8]. Human body silhouette is extracted from these equalization frames using segmentation GLCM segmentation technique [23] and normalised to into fixed cells. Further, the PCA technique [24] is used for dimension reduction, and classification of action classes is done through SVM linear classifier is used for action recognition. The Fig.1 depicts the proposed framework of our model.

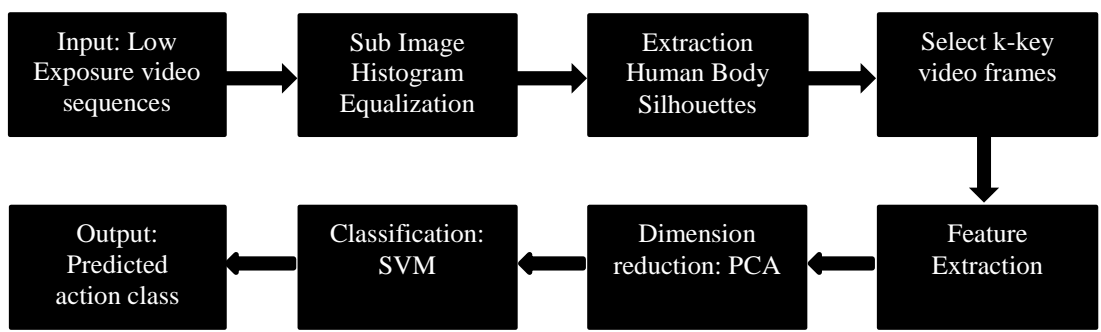

Fig. 1. Proposed framework for activity recognition.

### 3.1 Low contrast sampling of video sequences

It is a challenging task to track the object in low-quality videos due to less brightness, contrast and gamma factors. The brightness is one of the essential factors which discriminates the object from a background in video frames for better visual perception. A good contrast level distinguishes the one object from another in the video image and helps to track it. Similarly, the gamma factor is used to uniform linear contrast functions between the images and easy to recognized activity in videos.

In this work, the original video sequences downsampled into low-quality video frames, i.e. low contrast frames because of no such dataset available publically till now. In the video conversion process, no additional compress technique is used for output videos and retaining the original frames rate. The original video frames converted into subset of low quality frames which are having different combination of brightness and contrast such as: S1 (Brightness =-20%, Contrast = -20%), S2 (Brightness =-40%, Contrast = -40%), and S3 (Brightness =-50%, Contrast = -50%). These low-quality videos subsets represent meaningful feature extraction for action recognition. The Weizmann [25], KTH [26] and Ballet Movement [27] datasets were converted into low-quality video subsets for evaluation of our algorithm.

| Steps | Algorithm |
|---|---|
| 1 | Input: human actioned video sequences |
| 2 | Convert the original video sequences into low-quality sub-set S1, S2, and S3 video frames. |
| 3 | Compute the histogram of each subset frames. |
| 4 | Calculate the thresholding and exposing parameters. |
| 5 | Compute the clipping threshold histogram parameter. |
| 6 | Calculate sub image histogram from clipped based on threshold parameter. |
| 7 | Compute equalisation technique on all sub histograms. |
| 8 | Add the all subframes into a single frame for activity analysis. |
| 9 | Compute entropy images for each video sequences. |
| 10 | Apply a grey level $9 \times 9$ co-occurrence matrix to extract binary human silhouettes from entropy images. |
| 11 | Select the k-keyframes having significant energy compared to highest energy frames. |
| 12 | Divide the key frames into grid and cells. |
| 13 | Count the intensity value in each cells and put into feature vector |
| 14 | Output: Predicted Activity |

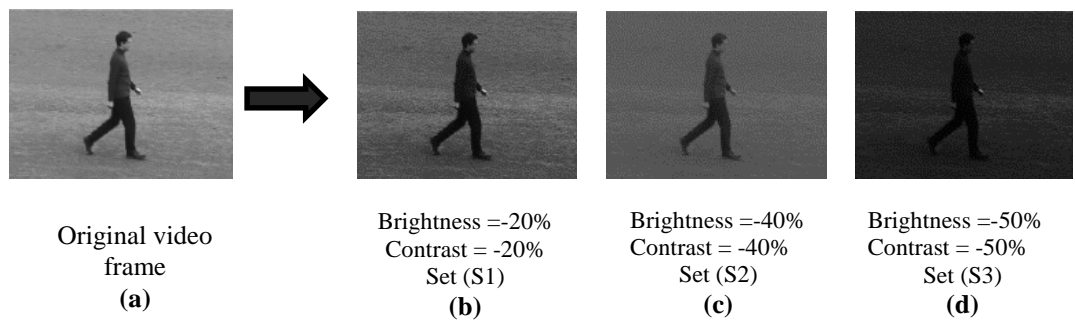

Original video frame
(a)

Brightness =-20%
Contrast = -20%
Set (S1)
(b)

Brightness =-40%
Contrast = -40%
Set (S2)
(c)

Brightness =-50%
Contrast = -50%
Set (S3)
(d)

Fig. 2 Low contrast video frames (b), (c), and (d) from original video frames (a).

## 3.2 Sub-Image Histogram Equalization Technique

The spatial domain image enhancement technique, based on sub-image histogram equalisation (ESIHE) showed effective performance on low exposure greyscale images while preserving the enhanced rate and entropy of the original images. Also, this approach provides a good enhancement rate for low-quality video frames.

This image enhancement technique is having mainly three steps, threshold calculation for exposure, histogram clipping, and subdivision equalisation.

Exposure threshold parameter is used to measure the intensity of the image or frame, and it can be defined as:

$$\mathcal{E}_t = \frac{1}{L} \frac{\sum_{k=1}^{L} h(k) k}{\sum_{k=1}^{L} h(k)} \qquad (1)$$

here $h(k)$ is a histogram of the image and L shows the grey level in input frames.
To divide the image into under and the overexposure region a parameter $£_a$ and it gives exposure value less or greater than L/2 for an image frame having dynamic value (0 to L) defined as:

$$£_a = L(1 - \mathcal{E}_t) \qquad (2)$$

To control the histogram enhancement rate histogram clipping is required which is calculated as:

$$T_c = \frac{1}{L} \sum_{k=1}^{L} h(k) \qquad (3)$$

$$h_c(k) = T_c \qquad for\ h(k) \geq T_c \qquad (4)$$

where $h(k)$ and $h_c(k)$ are the original and clipped histogram respectively.
The sub-image histogram equalization process consists of sub-images ranging from grey level (0 to L-1). If $\mathcal{P}_L(k)$ and $\mathcal{P}_U(k)$ are the probability density function (PDF) of these sub-images frames defined as:

$$\mathcal{P}_L(k) = \frac{h_c(k)}{\mathcal{N}_L} \qquad for\ 0 \leq k \leq £_a \qquad (5)$$

$$\mathcal{P}_U(k) = \frac{h_c(k)}{\mathcal{N}_U} \qquad for\ £_a + 1 \leq k \leq L - 1 \qquad (6)$$

where $\mathcal{N}_L$ and $\mathcal{N}_U$ are total number of pixels in sub-images.
The cumulative distribution function (CDFs) of these sub-images given,

$$\mathcal{C}_L(k) = \sum_{k=1}^{£_a} \mathcal{P}_L(k) \qquad (7)$$

$$\mathcal{C}_U(k) = \sum_{k=£_a+1}^{L-1} \mathcal{P}_U(k) \qquad (8)$$

The transfer function for histogram equalization is obtained from Eq.1 to Eq.6 defined as

$$\mathcal{F}_L = £_a \times \mathcal{C}_L \qquad (9)$$

$$\mathcal{F}_U = (£_a + 1) + (L - £_a + 1)\mathcal{C}_U \qquad (10)$$

Where $\mathcal{F}_L$ and $\mathcal{F}_U$ are the transfer function used for the sub-image histogram equalization respectively. The final output obtained from a combination of both transfer function of these sub-images into one complete image frame.

### *3.3 Extraction of Silhouette from texture information*

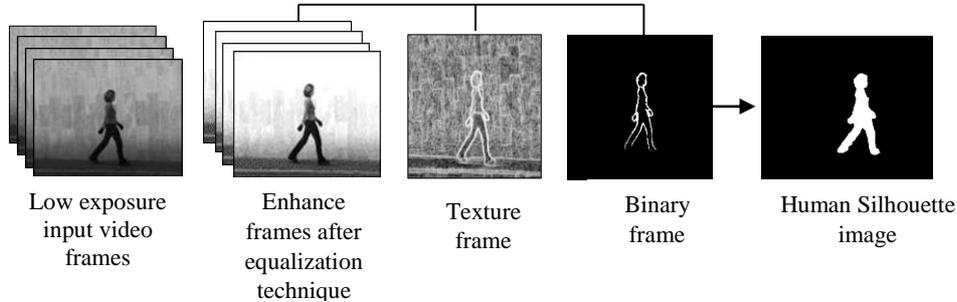

Low exposure input video frames | Enhance frames after equalization technique | Texture frame | Binary frame | Human Silhouette image

Fig. 3. Human silhouettes extraction using GLCM method [23] from enhance video sequences after enhancement.

Background subtraction is a fundamental step to recognized human activity from video sequences. We extracted human silhouette using grey level co-occurrence matrix (GLCM) method, which gives efficient results for video representation. The entropy of an image is the most critical parameter for classification of texture information in images. If entropy is higher than the details in the image is complex. It can be calculated as:

$$E = \sum_m \sum_n p(m,n) \log(p(m,n)) \qquad (11)$$

Where $p(m,n) = \frac{\mathcal{M}(m,n)}{\sum_{m,n} \mathcal{M}(m,n)}$ is PDF and $m$, $n$ are indices to the co-occurrence matrix $\mathcal{M}$.

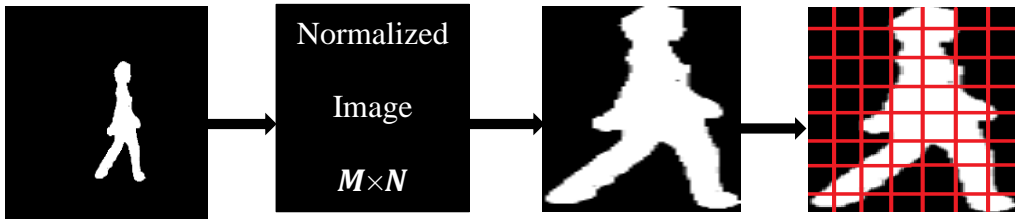

Fig 4. Cell formation from key frames

The different textural information in the image is represented using an entropy-based 9×9 neighbourhood filter. This filter matrix is converted into binary form with some thresholding and yields image with a white spot at different pixel locations. By comparing the contours, that having the largest area is considered as a human blob and is selected as a human silhouette.

### *3.4 Feature Extraction using k-key pose frames*

### *3.4.1 Extractions of k-key pose frames*

The robust features extraction is main objective of video sequences analysis that should be invariant to illumination, background changes, noise, and body poses etc. It can be observed that not all video frames consist of meaningful information about the object. Due to this, keyframes are extracted which contains maximum information as compared to other frames.

These keyframes are processed for feature extraction. The higher energy of k- key pose frames calculated as:

$$\mathcal{U}_t = \sum_i^M \sum_j^N \|\mathcal{I}_t(i,j)\|^2 \tag{12}$$

Such keyframes are arranged in time interval manner concerning higher energy frames. These frames are robust and invariant for discriminating various human actions. Further, these frames are divided into cells grids for calculating white pixel used in the feature extraction task.

*3.4.2 Feature computation and representation*

The total number of cells $T_c$ in the key frames given by

$$C_1, C_2, C_3 \ldots \ldots \ldots \ldots \ldots C_{T_c} \tag{13}$$

The number of white pixels $\rho_i$ in the binary silhouettes frames are counted as:

$$\mathfrak{w}_i = count\{C_i(x,y)\}, \quad where\ i = 0,1,2, \ldots \ldots \ldots \ldots T_c \tag{14}$$

The final feature matrix for video dataset having, number of videos of all action classes can be demonstrated as:

$$\mathcal{F}_\mathcal{V} = \begin{pmatrix} \mathcal{F}_1 = \mathfrak{w}_1,\mathfrak{w}_2,\mathfrak{w}_3,\ldots\mathfrak{w}_{T_c},\ \mathfrak{w}_1,\mathfrak{w}_2,\mathfrak{w}_3,\ldots\ldots\mathfrak{w}_1,\mathfrak{w}_2,\mathfrak{w}_3,\ldots\mathfrak{w}_{T_c} \\ \mathcal{F}_1 = \mathfrak{w}_1,\mathfrak{w}_2,\mathfrak{w}_3,\ldots\mathfrak{w}_{T_c},\ \mathfrak{w}_1,\mathfrak{w}_2,\mathfrak{w}_3,\ldots\ldots\mathfrak{w}_1,\mathfrak{w}_2,\mathfrak{w}_3,\ldots\mathfrak{w}_{T_c} \\ \vdots \\ \mathcal{F}_{\mathbb{V}_T} = \mathfrak{w}_1,\mathfrak{w}_2,\mathfrak{w}_3,\ldots\mathfrak{w}_{T_c},\ \mathfrak{w}_1,\mathfrak{w}_2,\mathfrak{w}_3,\ldots\ldots\mathfrak{w}_1,\mathfrak{w}_2,\mathfrak{w}_3,\ldots\mathfrak{w}_{T_c} \end{pmatrix} \tag{15}$$

Where, $\mathcal{F}_i = [\mathfrak{w}_1, \mathfrak{w}_2, \mathfrak{w}_3, \ldots \mathfrak{w}_{T_c}, \mathfrak{w}_1, \mathfrak{w}_2, \mathfrak{w}_3, \ldots \ldots \mathfrak{w}_1, \mathfrak{w}_2, \mathfrak{w}_3, \ldots \mathfrak{w}_{T_c}]$ feature vector for a single video of activity.

The dimension of the final feature vector is calculated as:

$$\mathcal{F}_\mathcal{V} = \mathbb{V}_T \times T_c \tag{16}$$

For speedy classification rate, the dimension of a final feature vector or $\mathcal{F}_\mathcal{V}$ is further reduced by popular dimension reduction technique Principle Component Analysis (PCA) [24]. Finally, the lower dimension feature set is classified by most famous machine learning technique Support Vector Machine (SVM) [28].

**4 Experimental Result and Analysis**

We evaluated our experiments on three well known standard action datasets Weizmann [25], KTH [26] and Ballet Movement [27]. These datasets are recorded with uniform background and controlled environment conditions.

*4.1 Weizmann Action Dataset*

This was the first human activity dataset recorded by Gorelick et al. [25] in 2005. It was very simple and most familiar dataset for action recognition. The videos are recorded with a fixed camera in a static background and fixed illumination environmental conditions. It consists of total 90 videos in which nine subjects are doing ten action classes: "jumping jack", "running", "jumping", "walking", "bending", "forward jump with both legs", "jumping forward with one leg", "one hand waving", "two hands waving", "sideways jumping", and "jumping at the same place". This dataset is recorded with low spatial frame resolution with 144×180 at a frame rate of 15fps. The video samples frames of this dataset are depicted in Fig.5.

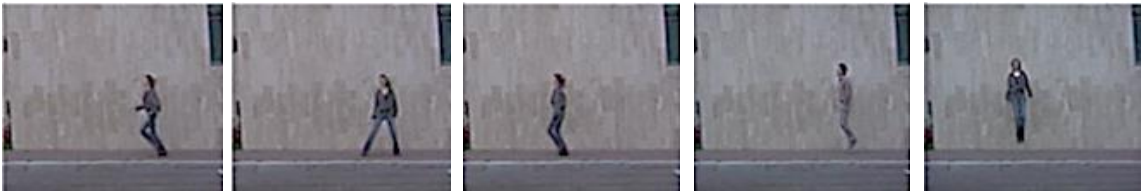

Fig. 5. Example frames of Weizmann action dataset [25]

*4.2 KTH Action Dataset*

KTH action [26] is found to be more difficult dataset than as compared to the Weizmann action due to changing environmental conditions. It is most frequently used dataset for human activity recognition in indoor as well as outdoor in changing illumination environment. It contains twenty-five subjects acting pre-defined six action classes: "jumping", "hand-clapping", "jogging", "running", "hand-waving", and "walking" in four different scenarios. All video sequences are recorded at the frame rate of 25 fps with a spatial resolution of 160x120 pixels. The sample frames from a video of KTH action dataset is depicted in Fig.6.

*4.3 Ballet Movement Dataset*

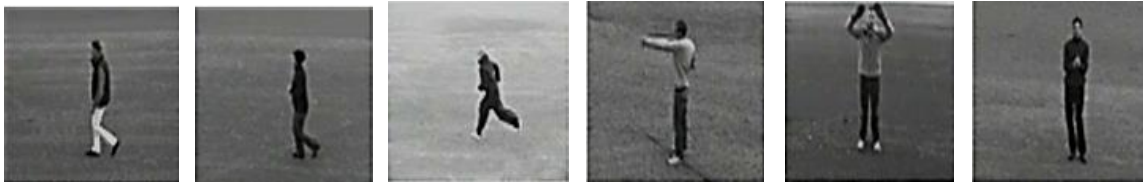
Fig. 6. Example frames of KTH action dataset [26]

The Ballet movement dataset [27] contains 8 ballet dancing activities performed by 3 dancers such as "standing hand opening", "standing still", "turning", "left-to-right hand opening", "leg swinging", "jumping", "hopping", and right-to-left hand opening. This dataset consists of 44 labelled video sequences. There are 3 actors, one woman and two men in which only one actor is performing in each video at a particular time. This dataset is challenging due to intra-class dissimilarity and inter-class similarity between different activities performed by dancers speed, apparels, spatial and temporal scales. The example sequences from video of Ballet dataset are shown in Fig.7.

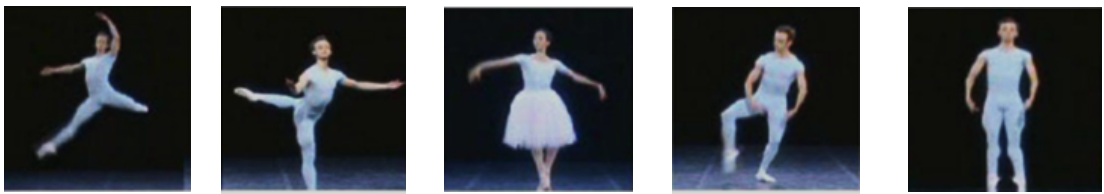
Fig. 7. Example frames of Ballet movement action dataset [27]

*4.4 Classification Results and Comparison*

The performance accuracy of our feature extractor is measured in terms of mean average precision (mAP), which is determined through action classification of datasets by SVM classifier in leave-one-out cross-validation (LOOCV) manner. The current implementation of the algorithm is run on an Intel® Core™ i5 3.2GHz processor with 8 GB memory card. The average recognition rate (ARR) is calculated using Eq. 17.

$$\text{Accuracy} = \frac{TP+TN}{TP+TN+FP+FP} \times 100\% \qquad (17)$$

where, TN, TP, FN, and FP are defined as true negative, true positive, false negative, and false positive respectively.

Table 1 shows the results of average recognition accuracy on three low-quality video sub-sets of standard action dataset. Further, the performance of these hybrid feature representations is comparable with our previous work [18] on original datasets. Tables 2, 3, and 4 depicted comparisons among the state-of-the-art approaches on these datasets.

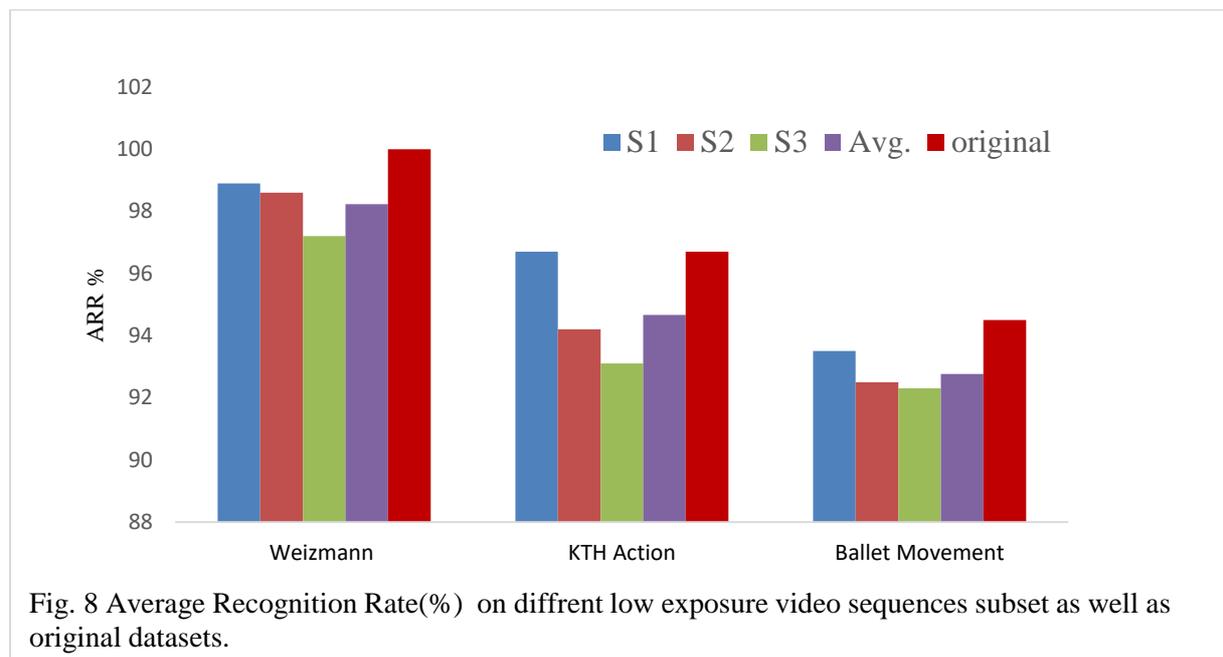

Fig. 8 Average Recognition Rate(%) on diffrent low exposure video sequences subset as well as original datasets.

**Table 1. ARR of Low-Quality Video Sequences of Standard Action Dataset**

| Methods | Weizmann | | | KTH Action | | | Ballet Movement | | |
|---|---|---|---|---|---|---|---|---|---|
| | S1 | S2 | S3 | S1 | S2 | S3 | S1 | S2 | S3 |
| **Proposed Method** | 98.9 | 98.6 | 97.2 | 96.7 | 94.2 | 93.1 | 93.5 | 92.5 | 92.3 |
| **ARA (%)** | **98.23** | | | **94.66** | | | **92.76** | | |

Table 2 shows the comparable accuracy of our model on low-quality video sequences as compared with similar approaches to Weizmann dataset.

**Table 2. Result comparison on Weizmann action dataset**

| Methods | Gorelick et al. [25] | Chaaraoui et al. [16] | Melfi et al. [29] | Our method |
|---|---|---|---|---|
| ARA (%) | 97.50 | 92.80 | 99.02 | **97.66** |

Table 1 shows the ARR of Weizmann dataset and cross-validation results of different actions. There are present challenges because of the high interclass similarity in key poses of running, walking, and jumping but our model outperforms in such kind of actions and each action discriminated with 97.66 % average recognition accuracy on three low-quality video sequences subset of the original dataset. There is slightly less confusion in the case of 'jack hand' and 'wave-2'due to similar activities.

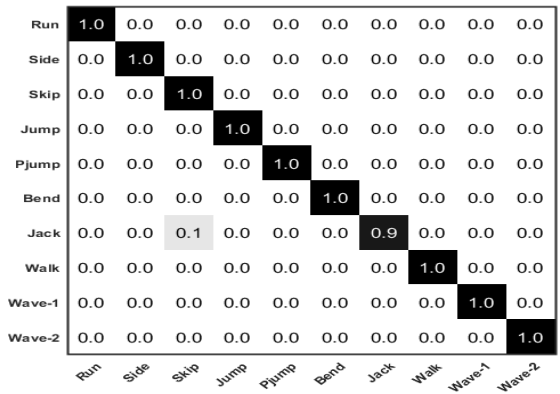
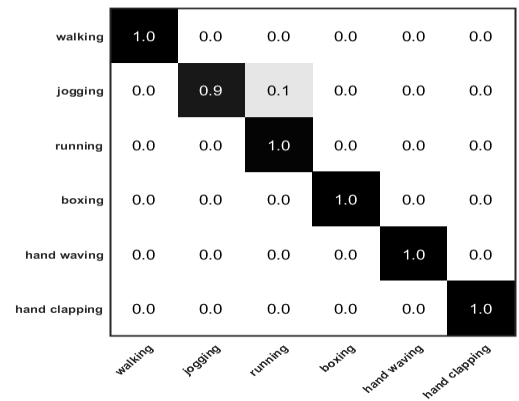

Fig. 9 (a)                                    Fig.9 (b)

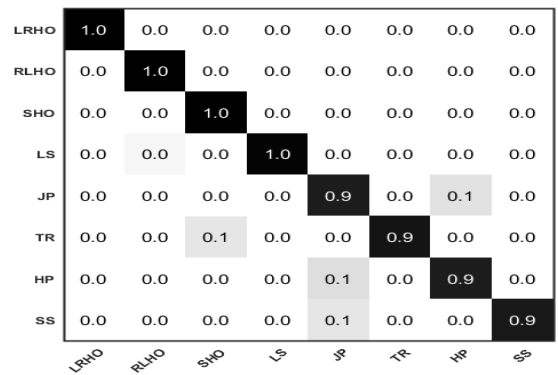

Fig.9 (c)

Fig. 9 Confusion matrix on (a) Weizmann (b) KTH (c) Ballet datasets.

Fig. 9 (a) shows the confusion matrix of Weizmann dataset and cross-validation results of different actions. There are present challenges because of the high interclass similarity in key poses of running, walking, and jumping but our model outperforms in such kind of actions and each action discriminated with 100 % recognition accuracy. There is slightly less confusion in the actions such as 'jack hand' and 'wave-2' due to similar activities.

**Table 3. Result comparison on KTH dataset**

| Methods | Saghafi and Rajan [30] | Melfi et al. [29] | Zheng et al. [31] | See and Rehman [19] | **Our method** |
| --- | --- | --- | --- | --- | --- |
| ARA (%) | 92.60 | 95.60 | 94.58 | 90.28 | **94.50** |

From Table 1, KTH dataset shows much satisfactorily results with average recognition accuracy 94.50 % on three low-quality video sequences subset of original dataset and much lesser misclassification is found only three classes: 'running', 'walking', and 'jogging' due to similar action. In Fig. 9 (b) the confusion matrix for KTH dataset shows much satisfactorily results with average recognition accuracy 98.70 %. Table 3 shows the best accuracy of our model on low-quality video sequences as compared with similar approaches to KTH dataset.

**Table 4. Result comparison on Ballet dataset**

| Methods | Fathi & Mori [27] | Wang and Mori [32] | Iosifidis et al. [33] | **Our Method** |
|---|---|---|---|---|
| ARA (%) | 51 | 91.30 | 91.10 | **92.36** |

The average recognition accuracy on three low-quality video sequences subset of an original dataset of 92.75 % computed through our method on Ballet dataset is shown in Table 1. The action recognition in this dataset is complicated due to clothing, gender, and size variations. Our feature extractor is insensitive to these variations and complexity of actions. It can be observed from the confusion matrix Fig. 9 (c) that there is little bit confused about of action pair such as 'hopping' and' jumping', 'leg swing' and 'Right to left-hand opening', 'turning right' and 'stand hand opening', and 'jumping and standing still' besides this our model outperform in comparison with the existing state-of-the-art methods.

## 5 Conclusion

In this work, we proposed a novel hybrid classification model based on the sub-image histogram equalisation technique which is more effective for enhancement on low exposure greylevel images. We have utilized the idea of such that a good quality enhances frames has higher entropy as well as high textural information as compared to low exposure frames. The human body silhouettes are extracted from these improve video frames for activity recognition. The key poses frames represent the spatiotemporal variations of human shapes. A robust feature vector is obtained from the grid and cells from these binary silhouettes and classified through a supervised machine learning technique, Support Vector Machine (SVM). In future, we can optimize the model for complex interaction and low-quality video surveillance applications.